%% file: root.tex
\documentclass[letterpaper, 10 pt, conference]{ieeeconf}  

\IEEEoverridecommandlockouts                              

\usepackage{amsmath} 
\usepackage{amssymb}  

\DeclareMathOperator{\trans}{trans}
\DeclareMathOperator{\rot}{rot}
\DeclareMathOperator{\inc}{inc}
\DeclareMathOperator{\Adj}{Adj}
\DeclareMathOperator{\diag}{diag}

\usepackage[utf8]{inputenc}
\usepackage[T1]{fontenc}
\DeclareUnicodeCharacter{2212}{-}
\usepackage{url}
\usepackage{subcaption}
\usepackage[inkscapearea=page]{svg}
\svgsetup{inkscapeformat=eps}
\usepackage{xcolor}
\usepackage{changes}
\usepackage[percent]{overpic}
\definechangesauthor[name=Wolfgang, color=blue]{WS}

\title{\LARGE \bf
  Towards Robust Monocular Visual Odometry for Flying Robots\\
  on Planetary Missions
}

\author{M. Wudenka$^{1,2}$, M. G. M{\"u}ller$^{1,3}$, N. Demmel$^{2}$, A. Wedler$^{1}$, R. Triebel$^{1,2}$, D. Cremers$^{2}$, W. St{\"u}rzl$^{1}$
  \thanks{$^{1}$
    Institute of Robotics and Mechatronics, German Aerospace Center (DLR)
. {\tt\small \{firstname.lastname\}@dlr.de}
        }
  \thanks{$^{2}$
    Computer Vision Group, Department of Informatics, Technical University of Munich, Germany %
        }
  \thanks{$^{3}$
    Autonomous Systems Lab, ETH Zurich, Switzerland%
        }
}

\newcommand\copyrighttext{%
  \footnotesize \textcopyright 2021 IEEE. Personal use of this material is permitted.
  Permission from IEEE must be obtained for all other uses, in any current or future
  media, including reprinting/republishing this material for advertising or promotional
  purposes, creating new collective works, for resale or redistribution to servers or
  lists, or reuse of any copyrighted component of this work in other works.
}
\newcommand\copyrightnotice{%
\begin{tikzpicture}[remember picture,overlay]
\node[anchor=south,yshift=10pt] at (current page.south) {\fbox{\parbox{\dimexpr\textwidth-\fboxsep-\fboxrule\relax}{\copyrighttext}}};
\end{tikzpicture}%
}

\begin{document}
\bstctlcite{IEEEexample:BSTcontrol}

\maketitle
\copyrightnotice
\thispagestyle{plain}
\pagestyle{plain}

\begin{abstract}


In the future, extraterrestrial expeditions will not only be conducted by rovers but also by flying robots. 
The technical demonstration drone Ingenuity, that just landed on Mars, will mark the beginning of a new era of exploration unhindered by terrain traversability.
Robust self-localization is crucial for that.
Cameras that are lightweight, cheap and information-rich sensors are already used to estimate the ego-motion of vehicles.
However, methods proven to work in man-made environments cannot simply be deployed on other planets.
The highly repetitive textures present in the wastelands of Mars pose a huge challenge to descriptor matching based approaches.

In this paper, we present an advanced robust monocular odometry algorithm that uses efficient optical flow tracking to obtain feature correspondences between images and a refined keyframe selection criterion. In contrast to most other approaches, our framework can also handle rotation-only motions that are particularly challenging for monocular odometry systems. Furthermore, we present a novel approach to estimate the current risk of scale drift based on a principal component analysis of the relative translation information matrix. This way we obtain an implicit measure of uncertainty.
We evaluate the validity of our approach on all sequences of a challenging real-world dataset captured in a Mars-like environment and show that it outperforms state-of-the-art approaches.
The source code is publicly available at: \\ \centerline{\url{https://github.com/DLR-RM/granite}}
\end{abstract}

\input{chapters/introduction.tex}

\input{chapters/related-work.tex}

\input{chapters/notation.tex}

\input{chapters/monocular-odometry.tex}

\input{chapters/uncertainty.tex}

\input{chapters/evaluation.tex}
\addtolength{\textheight}{-0.0cm}   

\input{chapters/conclusions.tex}






\section*{ACKNOWLEDGMENT}

This work was supported by the Helmholtz Association, project ARCHES ({\tt\small www.arches-projekt.de/en/}, contract number ZT-0033).

\bibliographystyle{IEEEtran}
\bibliography{IEEEabrv,literature}

\end{document}

%% file: chapters/introduction.tex

\section{INTRODUCTION}

\begin{figure}[!t]
    \vspace*{2.5mm}
    \begin{subfigure}[t]{\columnwidth}
        \includegraphics[width=\textwidth]{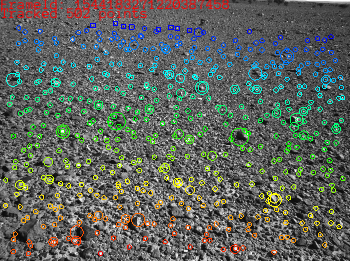}
        \caption{Landmarks projected onto the image with their estimated inverse distance color coded. Circles illustrate the feature patch size at highest resolution, resulting in larger circles for features initialized at lower resolutions (see Sec.~\ref{sec:mono-odometry}-A).}
    \end{subfigure}
    \vspace*{2.5mm}

    \begin{subfigure}[t]{\columnwidth}
        \includegraphics[width=\textwidth]{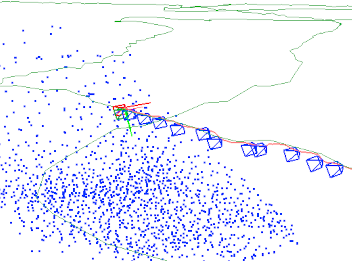}
        \caption{Estimated trajectory (red) and the ground truth trajectory based on GPS data (green) are shown, as well as current camera pose (red) and the keyframe camera poses (blue).}
    \end{subfigure}
    \caption{Our robust monocular odometry at work in a Mars-analogue environment (example from sequence D-3 of the MADMAX dataset~\cite{madmax}).}
    \label{fig:tracking-madmax}
\end{figure}

Since 1997, selected regions of Mars have successfully been explored by robotic vehicles, performing more and more tasks autonomously.
Essential for their autonomy is the ability of self-localization aided by visual odometry (VO), i.e. dead reckoning ego-motion from an image stream.
Robust and accurate state estimation becomes even more important for flying robots, since they have to work fully autonomously. This is the case for future mission concepts~\cite{arches} and also for the current NASA Mars 2020 mission and its flying robot Ingenuity.
The VO of Ingenuity is designed for limited computational resources and flight durations of about $90\,\mathrm{s}$ on flat terrain~\cite{ingenuity-vo}. 

Being able to robustly navigate a flying robot in any environment on Mars would enable fast exploration, independent of the traversability of the terrain.
Flying robots can act as scouts for rovers and also discover and explore areas where rovers cannot reach. 
Furthermore, flying swarms that increase the overall fault tolerance through redundancy are conceivable, although increasing the need for cheap and lightweight drones. 

In the last decades, academic research has brought up various different solution strategies to tackle the visual odometry problem. 
However, when evaluating recent state-of-the-art systems on datasets which resemble Mars-like environments such as the one recorded in the Morocco-Acquired Dataset of Mars-Analogue eXploration (MADMAX)~\cite{madmax}, we especially found that methods relying on descriptor-based feature matching lack robustness.
While direct methods seem promising, they still require significantly more computational resources and are overall not as mature.
They heavily rely on a motion model as initialization for the optimization algorithm.
In our experiments we found that indirect methods based on Kanade-Lucas-Tomasi (KLT) optical flow tracking~\cite{KLT} for data association balance robustness, accuracy and computational demands well.

For visual odometry, a monocular setup is especially challenging as the scene scale cannot be observed and thus, the gauge freedom has seven degrees; one more than that of a stereo setup. 
Hence, there is one more dimension where drift accumulates. 
Moreover, during pure rotation, no feature positions can be measured by triangulation. 
On the other hand, a single camera as main sensor is appealing as being lightweight and cheap. 
Furthermore, stereo setups effectively degrade to monocular when flying at high altitudes, as the ratio of stereo baseline and feature distance becomes small. 
In this paper, we therefore describe how we generalize the VO described in~\cite{basalt} to monocular setups. 
Moreover, we show that an enhanced feature-point and keyframe selection can improve the robustness on repetitively textured environments, as present in the planetary-like MADMAX dataset. 

Crucial to the robustness of the whole robotic system is knowledge of the uncertainties caused by each component. 
Consequently, we explore probabilistically motivated methods to estimate the current tracking quality and to detect the risk of scale drift. 
Such information is relevant for decision making and path planning. 

To summarize, our contributions are:
\begin{itemize}
    \item Generalization of the VO presented in \cite{basalt} to monocular setups (Sec.~\ref{sec:map-initialization}).
    \item Handling of rotation-only motion (Sec.~\ref{sec:points-infinity}). 
    \item Increased robustness in environments with highly repetitive textures as present in planetary exploration as well as to frames drops and large motions between frames by means of improved feature initialization (Sec.~\ref{sec:patch-scale}) and keyframe selection (Sec.~\ref{sec:keyframe-selection}).
    \item A novel probabilistic measure as indicator for scale drift (Sec.~\ref{sec:uncertainty}).
    \item Comparison of our system with other state of the art methods on a dataset relevant to planetary exploration (Sec.~\ref{sec:benchmarks}). 
\end{itemize}
Fig.~\ref{fig:tracking-madmax} shows an example of our visual odometry running on a sequence of the MADMAX dataset~\cite{madmax}.

%% file: chapters/related-work.tex

\section{RELATED WORK}

In the last decades, various systems for tracking a camera through VO were developed. 
They all differ in the chosen data association method, optimization strategy and supported sensor modalities: monocular, stereo and/or the fusion of additional sensors such as Inertial Measurement Units (IMUs). 
Additionally, some methods aim for global consistency and therefore are called Simultaneous Localization and Mapping (SLAM) systems.
While modern VO systems achieve very high accuracy through bundle adjustment,
they do not provide an easy to interpret measure of confidence.
Therefore, we split our related work discussion in two parts. First, we present methods relevant to camera tracking and research about uncertainty estimation later. 

\textbf{Monocular Odometry:} 
Similar to our approach, the visual-inertial odometry of the Mars Helicopter~\cite{ingenuity-vo} uses KLT tracking~\cite{KLT} for data association. 
Moreover, an Extended Kalman Filter (EKF) is used to process landmarks that are all assumed to lie on a flat ground plane.
Thus, it cannot fly over hilly terrain. 
The scene scale is obtained via an altimeter pointing nadir.
In~\cite{range-VIO} an experimental system is presented that relaxes the flat surface assumption. 
Since Parallel Tracking and Mapping (PTAM)~\cite{PTAM} has proven that windowed geometric bundle adjustment can run in real time on modern hardware, many new smoothing systems were presented reaching new levels of accuracy. 
One of the most elaborate frameworks is ORB-SLAM3~\cite{ORB-SLAM3}. It uses descriptor based matching and optimizes a local window of the covisibility graph, as well as a global pose graph in parallel. 
ORB-SLAM3 can handle monocular, monocular-inertial, stereo and stereo-inertial setups.
VINS-Mono~\cite{VINS-Mono} is the most similar system to ours. It uses KLT tracking and fixed-lag smoothing in the frontend to fuse visual and inertial measurements. 
The odometry described in~\cite{basalt}, named ``Basalt'', uses KLT tracking as well.
Basalt can handle stereo and stereo-inertial setups, but not monocular ones. 
Because of its capabilities and features, which we will further summarize in Sec.~\ref{sec:mono-odometry}, we base our approach on it. 
Although Basalt and VINS-Mono have a lot in common, according to Table I in \cite{basalt}, Basalt performs better than VINS-Fusion, the stereo version of VINS-Mono.
In the seminal work of~\cite{DSO}, a direct but sparse method was presented with noticeable performance in situations where indirect methods show problems, as in the presence of motion blur.
However, the unstable map initialization and the reliance on a motion model for initializing the optimization algorithm hinders a deployment in our domain.

\textbf{Uncertainty Estimation:} 
Methods, that use Gauss-New\-ton optimization, compute the Fisher information matrix. Even though it encodes the uncertainty in the current state estimate, it is not straight forward using it for decision making.
Kuo et al.~\cite{arbitrary-camera-slam} suggested to use negative entropy, a scalar value describing the tracking quality of the current frame to aid keyframe selection.
We take this formulation a step further and investigate the tracking quality of the current optimization window to detect the risk of scale drift.

%% file: chapters/notation.tex

\section{NOTATION}

\begin{figure*}
    \centering
    {\scriptsize\includesvg[width=\textwidth]{figures/flow-chart}}
    \caption{Flow chart that compares a) the original stereo-based Basalt-VO~\cite{basalt} with b) our monocular approach. Modifications and additions are highlighted in red.}
    \label{fig:flow-chart}
\end{figure*}

Throughout this paper, we express scalars as light lowercase letters $a$, vectors as bold lowercase letters $\boldsymbol{b}$ and matrices as bold uppercase letters $\boldsymbol{C}$. 
Elements of the Special Euclidean Group $SE(3)$ are denoted as $\boldsymbol{T}_i^j$ so that they transform a point in coordinate frame $i$ into $j$.
In that context index '$\mathrm{w}$' denotes the world reference frame. 
The matrix representation of such a rigid-body transformation is
\begin{equation}
    \boldsymbol{T} = \begin{bmatrix}
        \boldsymbol{R} & \boldsymbol{t} \\
        \boldsymbol{0}_{1,3} & 1
    \end{bmatrix}.
\end{equation}

The operators $\trans(\boldsymbol{T}) = \boldsymbol{t}$ and $\rot(\boldsymbol{T}) = \boldsymbol{R}$ extract the translational part $\boldsymbol{t} \in \mathbb{R}^3$ and rotational part $\boldsymbol{R} \in SO(3)$ from the transformation, respectively.
The operator $[\cdot]_\times$ converts the tangent vectors $\boldsymbol{w} \in \mathbb{R}^3$ (Rodrigues vectors) to elements of the Lie algebra $\mathfrak{so}(3)$:
\begin{equation}
    [\boldsymbol{w}]_\times = \begin{bmatrix}
        0 & -w_3 & w_2 \\
        w_3 & 0 & -w_1 \\
        -w_2 & w_1 & 0 \\
    \end{bmatrix}.
\end{equation}

The exponential map that converts elements from Lie algebra to the manifold $SO(3)$ is denoted $\exp : \mathfrak{so}(3) \rightarrow SO(3)$. 

Similarly to Rodrigues vectors, the elements of the Lie algebra $\mathfrak{se}(3)$ can be expressed as so called twist vectors, i.e. pose increments $\boldsymbol{\xi}\in\mathbb{R}^{6}$.
To convert a twist vector from the tangent space around one transformation to the tangent space of another one the adjoint action is defined as
\begin{equation}
    \Adj_{\boldsymbol{T}} = \begin{bmatrix}
        \boldsymbol{R} & [\boldsymbol{t}]_\times \boldsymbol{R} \\
        \boldsymbol{0}_{3,3} & \boldsymbol{R}
    \end{bmatrix}.
\end{equation}
\vspace*{1mm}

%% file: chapters/monocular-odometry.tex

\section{MONOCULAR ODOMETRY}\label{sec:mono-odometry}

\begin{table*}
  \centering
  \caption{Tracking failure count per MADMAX dataset sequence~\cite{madmax}. $\dagger$: FAST corners are only searched for in the finest level; $\ddag$: the current frame is made keyframe.}
  \label{tab:eval-mono-odo}
  \begin{tabular}{| c | c c c c c c c c c c c c c c c c c c |}
      \hline
      & A-0 & A-1 & A-2 & A-3 & A-4 & A-5 & A-6 & B-0 & B-1 & B-2 & B-3 & B-4 & B-5 & B-6 & B-7 & C-0 & C-1 & C-2 \\
      \hline
      Ours & 1 & 2 & 1 & 1 & 6 & 0 & 8 & 5 & 0 & 5 & 0 & 0 & 0 & 0 & 0 & 0 & 3 & 4\\
      Ours $\dagger$ & 1 & 2 & 2 & 1 & 7 & 1 & 10 & 5 & 0 & 4 & 1 & 0 & 0 & 0 & 0 & 3 & 3 & 3 \\
      Ours $\ddag$ & 13 & 4 & 1 & 3 & 60 & 7 & 99 & 62 & 14 & 7 & 0 & 0 & 1 & 1 & 0 & 12 & 7 & 10 \\
      \hline\hline
      & D-0 & D-1 & D-2 & D-3 & D-4 & E-0 & E-1 & E-2 & F-0 & F-1 & F-2 & F-3 & F-4 & F-5 & G-0 & G-1 & G-2 & H-0 \\
      \hline
      Ours & 2 & 3 & 2 & 8 & 7 & 4 & 5 & 2 & 0 & 0 & 2 & 1 & 3 & 2 & 1 & 0 & 8 & 0 \\
      Ours $\dagger$ & 2 & 3 & 2 & 8 & 10 & 5 & 6 & 2 & 0 & 0 & 2 & 1 & 3 & 2 & 2 & 0 & 8 & 1 \\
      Ours $\ddag$ & 2 & 17 & 4 & 12 & 18 & 11 & 19 & 2 & 2 & 10 & 4 & 6 & 26 & 10 & 10 & 20 & 36 & 1 \\
      \hline
  \end{tabular}
\end{table*}

Basalt, as described in~\cite{basalt} is a stereo and stereo-inertial odometry that minimizes the geometric reprojection error.
Utilizing sparse KLT optical flow tracking, it estimates the $SE(2)$ transform of FAST corners~\cite{FAST} from the previous left camera image to the current and from the current left camera image to the image of the right camera. 
To achieve robustness at the occurrence of large optical flow, a pyramidal approach is used, tracking feature patches from the coarsest to finest level.
For outlier detection, every feature is tracked forward and backward. 
Only features that return to the starting point are considered valid. 
To estimate the current camera pose a fixed-lag smoothing~\cite{fixed-lag} strategy on a sparse set of keyframes is applied. 
Two feature points $\in \mathcal{P}$ in two keyframes $\in \mathrm{obs}(i)$ connected by KLT-tracking are considered observations of the same landmark $i$ at pixel coordinates $\boldsymbol{z}_{it}$. 
Every landmark is hosted by one keyframe $h(i)$ and expressed in the corresponding coordinate frame. 
Landmarks are parameterized as bearing vectors and inverse distances $d$.
An initial inverse distance is determined by triangulation. 
Keyframe poses, landmark bearing vectors and inverse distances are optimized jointly in a sliding window (fixed-lag smoothing).
Old keyframes are marginalized into the prior $E_{\mathrm{m}}$ to keep the problem size bounded. 
Thus, Basalt optimizes the cost function 
\begin{align}
  E &= E_{\mathrm{reproj}} + E_{\mathrm{m}}\label{eq:basaltCost} \\
  E_{\mathrm{reproj}} &= \sum_{\substack{i \in \mathcal{P} \\ t \in \mathrm{obs}(i)}} \boldsymbol{r}_{it}^\top \Sigma_{it}^{-1} \boldsymbol{r}_{it} \\
  \boldsymbol{r}_{it} &= \boldsymbol{\pi} \left( \boldsymbol{T}_{h(i)}^t \boldsymbol{l}_i \right) - \boldsymbol{z}_{it}\label{eq:reprErr} \text{.}
\end{align}
$\boldsymbol{T}_{h(i)}^t$ is the relative pose from host keyframe to observing keyframe, $\boldsymbol{l}_i$ the landmark position in front of the host frame and $\boldsymbol{\pi}$ the camera projection function.
Basalt also features a globally consistent mapping layer, which is not considered in this paper.

We chose to base our work on Basalt for the following reasons:
\begin{itemize}
  \item Robustness: Basalt shows robust tracking results on the MADMAX dataset. The robustness of KLT-tracking in space exploration is also demonstrated by~\cite{ingenuity-vo}.
  \item Computational efficiency: Sparse optical flow tracking and fixed-lag smoothing well balance performance and computational demands.
  \item General camera models: Basalt uses bearing vectors. So it can be used with arbitrary camera models, allowing us to use wide-angle lenses.
  \item Code quality: Basalt is well documented and unit tested.
\end{itemize}

However, Basalt does not meet all of our requirements for lightweight flying robots for space exploration:
\begin{itemize}
  \item Monocular systems are not supported.
  \item Rotation-only motion cannot be tracked.
  \item The keyframe selection heuristics are hard to tune and hinder robust operation.
  \item No uncertainty measure is provided.
\end{itemize}

In the following sections, we detail our improvements to Basalt.
Firstly, we explain and evaluate how our method increases robustness by improving the feature point detection (Sec.~\ref{sec:patch-scale}).
Secondly, we introduce tracking of rotation-only motion (Sec.~\ref{sec:points-infinity}) into the pipeline.
Subsequently, we present our map initialization for monocular configuration (Sec.~\ref{sec:map-initialization}) and robustified keyframe selection (Sec.~\ref{sec:keyframe-selection}).
The differences between Basalt and our system are highlighted in Fig.~\ref{fig:flow-chart}.

\subsection{Feature Detection at Different Scale Levels}\label{sec:patch-scale}

We found that tracking becomes unstable with high image resolutions when features are only detected at the finest level, especially during events causing high optical flow, e.g., rapid motion or frame drops.
Using more image pyramid levels makes the system invariant to different image resolutions and scene object scales.
Given an image pyramid with $n$ levels, we not only initialize feature patches for FAST corners at level $0$ (highest resolution), but up to level $n-2$. 
In subsequent images, all patches are then tracked from the coarsest to their initialization level. 
The detection of corner features at different levels results in a much higher usable feature count.
Features found in coarser levels are more distinctive in the coarsest levels, which helps when large optical flow occurs. 
To account for the different tracking resolutions, we weight the reprojection errors according to the feature initialization level $c$ with
\begin{equation}
    w = \frac{1}{2^c}.
\end{equation}

We evaluate this modification by counting the number of restarts after tracking failures for the different MADMAX~\cite{madmax} sequences, presented in Table~\ref{tab:eval-mono-odo}.
The numbers in the first row (``Ours'') are usually smaller or equal to those in the second (``Ours~$\dagger$''), which indicates that initializing features on all resolution levels leads to a reduction of tracking failures.
Note that the numbers for sequences~B-2 and~C-2 are not meaningful as tracking failures occur only at the end of the sequence where people walk around, covering a large part of the camera's field of view. 

\subsection{Pure Rotation Tracking}\label{sec:points-infinity}

\begin{figure*}[!tb]
  \begin{subfigure}[t]{0.48\textwidth}
      \centering
      \includegraphics[width=\textwidth]{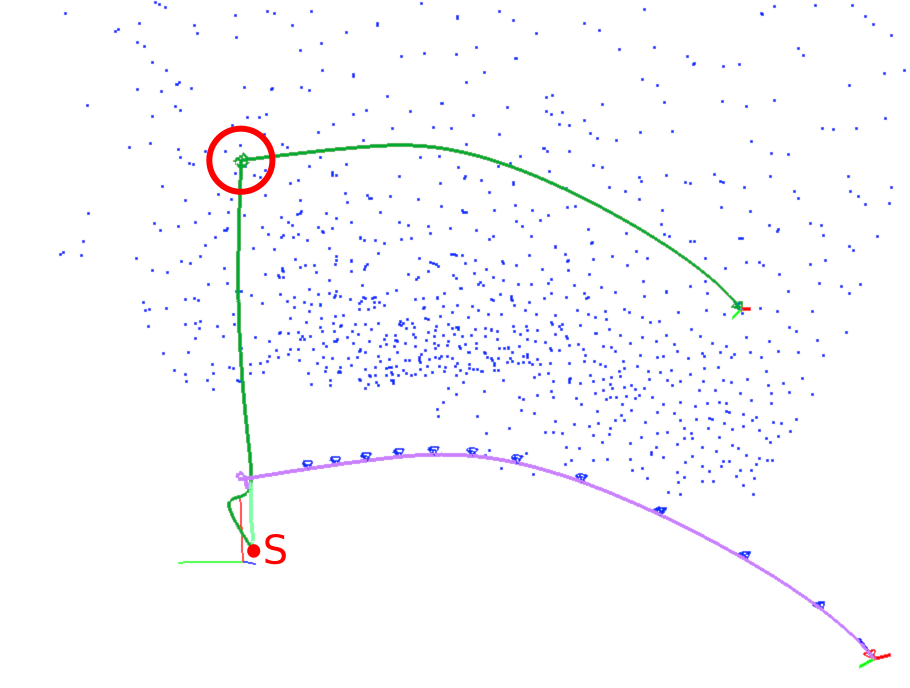}
      \caption{Two sub-maps (turquoise and purple) with different scales separated due to large rotation-only motion (red circle).}
  \end{subfigure}%
  \hfill%
  \begin{subfigure}[t]{0.48\textwidth}
      \centering
      \begin{overpic}[width=\textwidth]{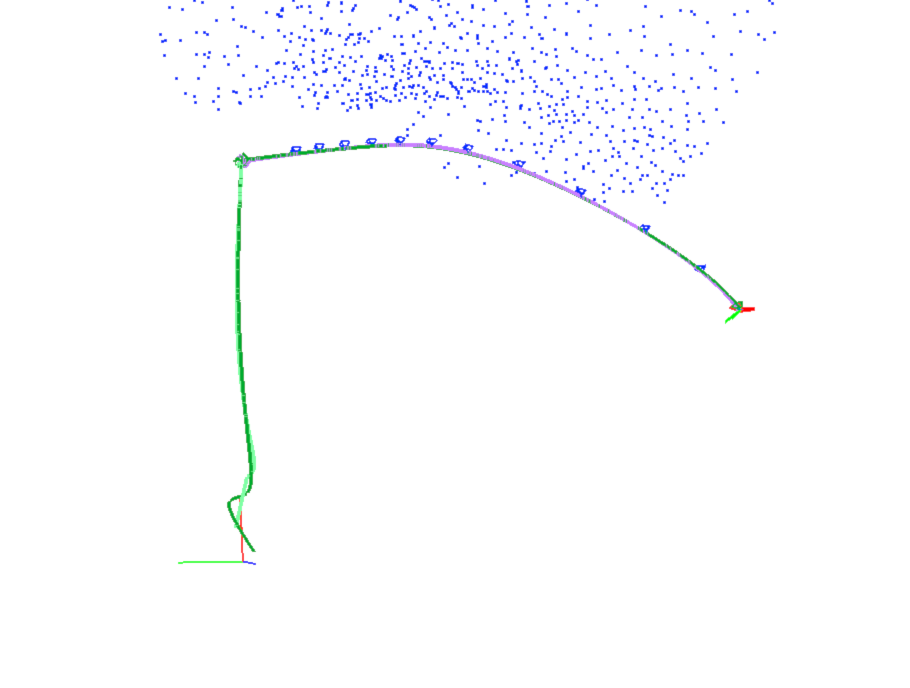}
        \put (46,15) {
          \definecolor{green}{rgb}{0, 0.67, 0.18}
          \definecolor{turquoise}{rgb}{0.5, 1, 0.65}
          \definecolor{purple}{rgb}{0.78, 0.5, 0.99}
          \definecolor{blue}{rgb}{0, 0.2, 1}
          \definecolor{palered}{rgb}{0.93, 0.23, 0.11}
          \begin{tabular}{r@{\hskip 0.5ex}l@{\hskip 3ex}r}
            {\textcolor{green}{\rule{1ex}{1ex}}} & ground truth \\
            {\textcolor{turquoise}{\rule{1ex}{1ex}}}{\textcolor{purple}{\rule{1ex}{1ex}}} & estimated trajectory \\
            \textbf{\textcolor{red}S} & start \\
            {\textcolor{blue}{\rule{1ex}{1ex}}} & landmarks, keyframes \\
            {\textcolor{palered}{\rule{1ex}{1ex}}} & current frame
          \end{tabular}
        }
       \end{overpic}
      \caption{The two sub-maps after separate $\mathrm{Sim}(3)$-alignment to ground truth.}
  \end{subfigure}

  \caption{After rotation-only motion a new sub-map is initialized, leading to a new scene scale. However, the different sections could be linked by a SLAM system. Landmarks, keyframes and current frame are shown only for the last optimization window of the second sub-map.}
  \label{fig:pure-rotation}
\end{figure*}

We explicitly allow landmarks to have an inverse distance $d$ of $0$ ("at infinity").
Such points contribute only to the estimation of rotation but not of translation.
Contrary to Basalt, we add a small user-defined prior tying the relative translation between keyframes connected by landmarks with $d=0$ to $\boldsymbol{0}$, avoiding an undetermined system.
A landmark becomes a ``point at infinity'' in two ways: The first possibility is that during initial triangulation no baseline between two frames which observe the landmark is larger than a threshold; the landmark is then inserted into the map with $d = 0$.
The second possibility can occur during bundle adjustment. Optimizing the inverse distance $d$ of a landmark is essentially a constrained optimization problem with $d \geq 0$.
During unconstrained Gauss-Newton (GN) optimization, it can happen that $d < 0$. This is solved by a projection into the valid region $d \leftarrow \max(0, d)$ after each GN update step, resulting in $d=0$. 

Allowing landmarks with $d = 0$ enables us to implicitly track rotation-only camera motion without the need for extra panorama maps as in~\cite{pure-rotation}. 
This feature is especially important for MAVs that want to take panorama overviews from a high altitude. 
Similarly, we can track pure rotations before a map is initialized. 
However, if a monocular odometry looses all points with $d > 0$, the following relative pose map initialization (see Sec.~\ref{sec:map-initialization}) determines a new scene scale. 
The resulting two sub-maps can be scale-aligned by a higher level system, as for example~\cite{ORBSLAM-atlas}.
However, this is out of the scope of this paper.
A scale-alignment using ground-truth data is visualized in Fig.~\ref{fig:pure-rotation}.
Every time a new observation of a point with $d = 0$ was made, we perform a new triangulation attempt. 

\subsection{Map Initialization}\label{sec:map-initialization}

In case there is no landmark with $d > 0$, the relative translation between two frames cannot be measured. 
This happens, when the map does not contain any landmarks (after system startup or reset due to tracking failure) or when the map only holds landmarks with $d = 0$ (after pure rotational motion around the camera center). 
Therefore, whenever less than five observations of landmarks with $d > 0$ are found in the current image, we attempt a relative pose map initialization. 
Similarly to~\cite{ORB-SLAM}, for every previous keyframe we generate relative motion hypotheses to the current frame with homography~\cite{homography} and five-point algorithm~\cite{five-point} in a parallelized RANSAC scheme~\cite{RANSAC}. 
Note, that after system startup our keyframe selection strategy (Sec.~\ref{sec:keyframe-selection}) ensures that the first frame is assigned as a keyframe. 
The best hypothesis in terms of reprojection errors is compared to the rotation-only estimate. 
If it has more inliers and a smaller sum of reprojection errors we attempt a map initialization: 
First, we scale the translation such that the average distance of the triangulated points is $\bar{\rho}$, arbitrary but fixed. 
Then, as proposed in~\cite{pure-rotation}, we approximate the parallax $\alpha$ of the scene as
\begin{equation}
  \alpha = 2 \arctan \left( \frac{t}{2\bar{\rho}} \right)
\end{equation}
where $t$ is the relative distance of the two frames used for initialization. 
If the parallax is higher than a threshold $\bar{\alpha}$, we consider the initialization attempt successful. 
Throughout our evaluation we found that $\bar{\alpha} = 5^\circ$ yields robust results.
To fix the scene scale we add a new error term $E_{\mathrm{scale fix}}$ to the optimization problem as defined in Eq.~(\ref{eq:basaltCost}): 
\begin{equation}
  E_{\mathrm{scale fix}} = w_{\mathrm{scale fix}} \left( | \trans((\boldsymbol{T}_j^{\mathrm{w}})^{-1} \boldsymbol{T}_i^{\mathrm{w}}) | - t \right)^2 \text{.}
\end{equation}
Here, $w_{\mathrm{scale fix}}$ is a user-defined weight factor.
$\boldsymbol{T}_j^{\mathrm{w}}$ and $\boldsymbol{T}_i^{\mathrm{w}}$ are the absolute poses of the frames used for map initialization.
This constraint is added to the marginalization prior as soon as either $i$ or $j$ is removed from the current optimization window.

\subsection{Keyframe Selection}\label{sec:keyframe-selection}

The keyframe selection of Basalt is based on heuristics. 
The user has to define a) a minimum number of frames between two keyframes and b) a threshold percentage for how many features in the current frame should successfully be associated to landmarks.
We found that a lot of fine tuning is necessary to adapt those values to new setups and other datasets. 
In comparison, the approach presented in~\cite{arbitrary-camera-slam} generalized better to other settings:
After a few iterations of non-linear optimization, we evaluate the negative entropy $E$ of the pose of the new frame $\boldsymbol{T}_n$ as
\begin{equation}
  E(\boldsymbol{T}_n) = \ln(|\boldsymbol{I}_{{\boldsymbol{T}}_n}|)
\end{equation}
using the Fisher information matrix $\boldsymbol{I}_{{\boldsymbol{T}}_n}\in \mathbb{R}^{6\times 6}$.
If $E(\boldsymbol{T}_n)$ drops below a certain threshold percentage of the running average of the negative entropy since the last keyframe insertion, a new keyframe is added.
Contrary to~\cite{arbitrary-camera-slam}, we do not make the current but the previous frame a keyframe. 
Typical cases where significant reduction of negative entropy can be observed are frame drops and rapid motions.
Those represent worst case scenarios for optical flow tracking as feature tracks might be torn apart.
Therefore, it is crucial to the robustness of a VO that every surviving feature track is harnessed.
We achieve this by making the frame before the negative entropy dropped a keyframe and thus, adding -- by means of triangulation -- new landmarks to the map.
The significant improvement due to this design decision can be seen in Table~\ref{tab:eval-mono-odo} when comparing the first row (``Ours'') with ``Ours~$\ddag$'' (third row).
``Ours~$\ddag$'' uses, as proposed in~\cite{arbitrary-camera-slam}, the frame for which the drop in negative entropy was detected as keyframe, resulting in significantly more tracking failures.

%% file: chapters/uncertainty.tex

\section{ESTIMATING SCALE DRIFT}\label{sec:uncertainty}

\begin{figure*}[!tb]
  \centering
  {\footnotesize
  \includesvg[width=\textwidth]{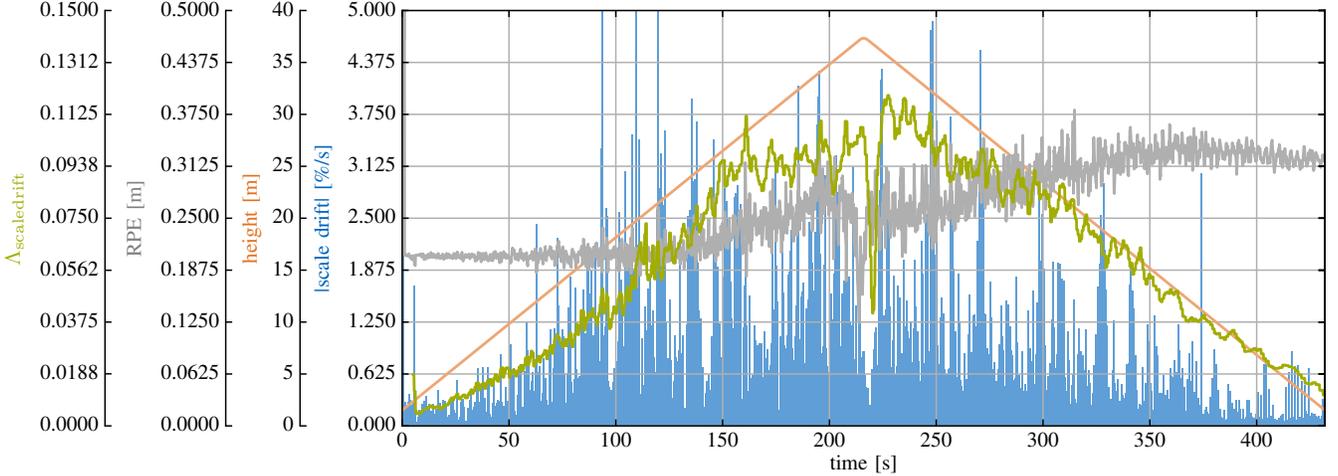}}
  \caption{Evaluation of scale drift and uncertainty estimation of our monocular method on a synthetic sequence. The virtual MAV first ascends and later descends (orange) in a spiral. The Relative Pose Error (RPE) (gray) is evaluated with a $\Delta t = 0.5\,\mathrm{s}$ and by $\mathrm{Sim}(3)$-aligning the first 100 frames. After map initialization (at $\sim 5\,\mathrm{s}$), $\Lambda_{\mathrm{scale drift}}$ (green) shows to be a good indicator for the actual relative scale drift (blue), computed by comparing the estimated trajectory to the perfect ground truth.}
  \label{fig:mono-neg-entropy}
\end{figure*}

\begin{figure}
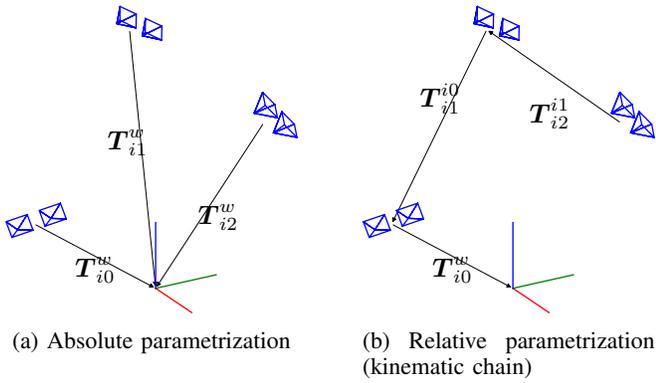

  \begin{subfigure}[t]{0.45\columnwidth}
      \includesvg[width=\textwidth]{figures/abs-parameterization}
      \caption{Absolute parametrization}
  \end{subfigure}\hfill%
  \begin{subfigure}[t]{0.45\columnwidth}
      \includesvg[width=\textwidth]{figures/rel-parameterization}
      \caption{Relative parametrization (kinematic chain)}
  \end{subfigure}
  \caption{Different ways of keyframe pose parametrization.}
  \label{fig:abs-rel-parametrization}
\end{figure}

Similar to the probabilistic keyframe selection criterion of~\cite{arbitrary-camera-slam}, we found that investigating the relative translation information matrix of the current optimization window leads to a good indicator for scale drift.
Since visual odometry is essentially a parameter estimation problem, we have measurements $\boldsymbol{z}$, the true state $\boldsymbol{x}$ and a measurement function $h(\hat{\boldsymbol{x}}) = \hat{\boldsymbol{z}}$ that relates the estimated state $\hat{\boldsymbol{x}}$ to the expected measurement $\hat{\boldsymbol{z}}$. 
In order to find $\hat{\boldsymbol{x}}$ that is as close to the unknown $\boldsymbol{x}$ as possible, we minimize a non-linear least squares problem with the residuals $\boldsymbol{r}(\hat{\boldsymbol{x}}) = \boldsymbol{h}(\hat{\boldsymbol{x}}) - \boldsymbol{z}$. 
Assuming that the measurements are disturbed by white noise with covariance $\boldsymbol{\Sigma}_{\boldsymbol{z}}$, the covariance of the current state can be obtained through 
\begin{equation}
  \boldsymbol{\Sigma}_{\hat{\boldsymbol{x}}} \approx \left( \boldsymbol{J}_{\boldsymbol{h}} ^\top \boldsymbol{\Sigma}_{\boldsymbol{z}}^{-1} \boldsymbol{J}_{\boldsymbol{h}} \right)^{-1} 
\end{equation}
where $\boldsymbol{J}_{\boldsymbol{h}}$ is the Jacobian of $\boldsymbol{h}$ evaluated at $\hat{\boldsymbol{x}}$. 
The inverse of the covariance $\boldsymbol{\Sigma}_{\hat{\boldsymbol{x}}}$ is also known as Fisher information $\boldsymbol{I}_{\hat{\boldsymbol{x}}} = \boldsymbol{\Sigma}_{\hat{\boldsymbol{x}}}^{-1}$.
Note that $\boldsymbol{J}_{\boldsymbol{h}} = \boldsymbol{J}_{\boldsymbol{r}}$ and therefore the matrix $\boldsymbol{J}_{\boldsymbol{h}}^\top \boldsymbol{\Sigma}_{\boldsymbol{z}}^{-1} \boldsymbol{J}_{\boldsymbol{h}}$ is the same as the $\boldsymbol{H}$ matrix computed in a standard Gauss-Newton optimization with update step $\Delta \hat{\boldsymbol{x}}$:
\begin{align}
  \Delta \hat{\boldsymbol{x}} &= - \left( \boldsymbol{J}_{\boldsymbol{h}}^\top \boldsymbol{\Sigma}_{\boldsymbol{z}}^{-1} \boldsymbol{J}_{\boldsymbol{h}} \right)^{-1}  \boldsymbol{J}_{\boldsymbol{h}}^\top  \boldsymbol{r}(\hat{\boldsymbol{x}}) \\
  &= - \boldsymbol{H}^{-1}  \boldsymbol{b}.
\end{align}

\begin{figure}[!tb]
  \begin{subfigure}[t]{0.48\columnwidth}
      \includegraphics[width=\textwidth]{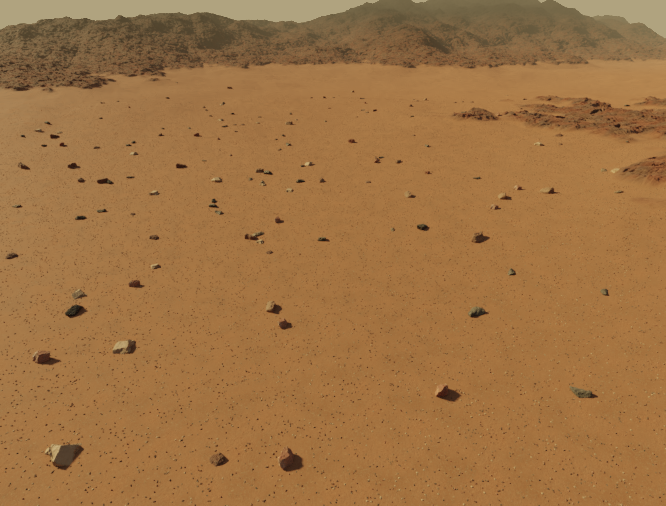}
  \end{subfigure}\hfill%
  \begin{subfigure}[t]{0.48\columnwidth}
      \includegraphics[width=\textwidth]{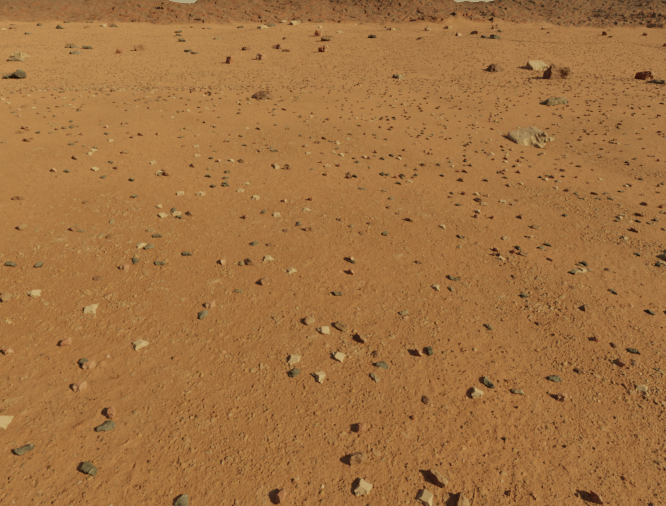}
  \end{subfigure}
  \caption{Our photo realistic recreation of the Martian surface using Blender~({\tt\small www.blender.org}). It allowed us to synthetically create arbitrary test sequences with perfect ground truth.}
  \label{fig:simulation-environment}
\end{figure}

\begin{figure*}[!tb]
  \centering
  {\scriptsize\includesvg[width=\textwidth]{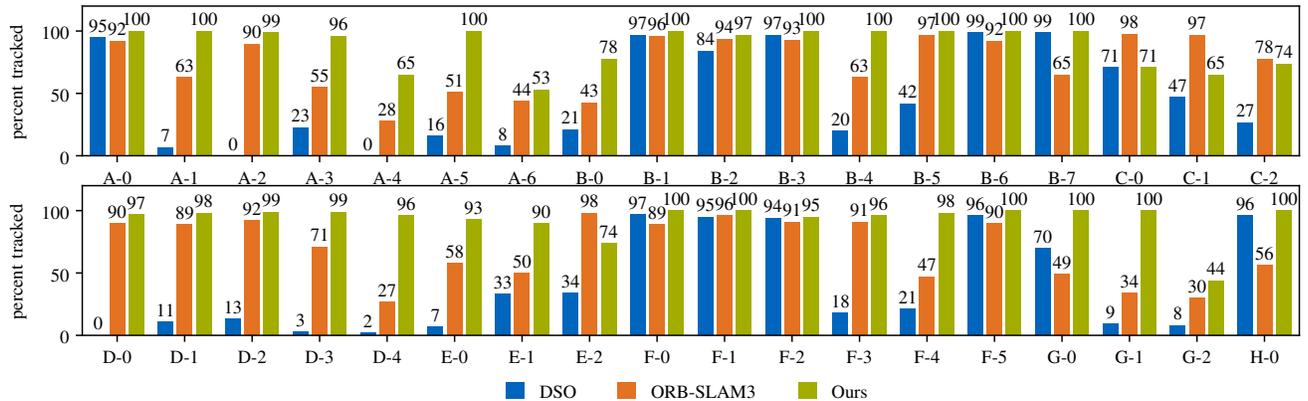}}
  \caption{Maximum relative sequence length that could be tracked on the MADMAX dataset~\cite{madmax}. The best value out of three runs is taken. Note that, in contrast to DSO, ORB-SLAM3 and our approach can restart after a failure. For those the longest continuously tracked sequence is taken. Numbers are rounded to nearest integer.}
  \label{fig:madmax-percent-tracked}
\end{figure*}

To obtain the Fisher information of the relative translations between keyframe states we reparametrize the problem as kinematic chain (Fig. \ref{fig:abs-rel-parametrization}). 
So the state $\boldsymbol{x}$ consists of $n$ relative keyframe poses $\begin{bmatrix}
    \boldsymbol{T}_0^1 & \boldsymbol{T}_1^2 & ... & \boldsymbol{T}_{n-1}^n
\end{bmatrix}$ and $m$ landmark positions $\begin{bmatrix}
    \boldsymbol{l}_0 & \boldsymbol{l}_1 & ... & \boldsymbol{l}_{m-1}
\end{bmatrix}$.
The relative pose in Eq.~(\ref{eq:reprErr}) is defined as $\boldsymbol{T}_{h(i)}^t = ({\boldsymbol{T}_t^\mathrm{w}})^{-1} \boldsymbol{T}_{h(i)}^\mathrm{w}$.
However, for the kinematic chain formulation this becomes $\boldsymbol{T}_{h(i)}^t = \boldsymbol{T}_{b_1}^t \boldsymbol{T}_{b_2}^{b_1} ... \boldsymbol{T}_{h(i)}^{b_n}$ or $\boldsymbol{T}_{h(i)}^t = ({\boldsymbol{T}_{t}^{b_1}})^{-1} ({\boldsymbol{T}_{b_1}^{b_2}})^{-1} ... ({\boldsymbol{T}_{b_n}^{h(i)}})^{-1}$ (depending on the traversal direction) for all keyframes $b_i$ that are in between the host and target frame. 
The Jacobian $\boldsymbol{J}_{\boldsymbol{r}_{it}}$ can then be obtained by taking the derivatives of $\boldsymbol{\pi}$ and $\boldsymbol{T}_{h(i)}^t$ and applying the chain rule.
Since we are using the same decoupled left increment as~\cite{basalt}, defined as
\begin{equation}
  \inc(\boldsymbol{T}, \boldsymbol{\xi} = \begin{bmatrix} \boldsymbol{v} \\ \boldsymbol{w} \end{bmatrix}) = \begin{bmatrix}
    \exp([\boldsymbol{w}]_{\times}) \boldsymbol{R} & \boldsymbol{t} + \boldsymbol{v} \\
    \boldsymbol{0}_{1,3} & 1
  \end{bmatrix}
\end{equation}
the partial derivatives of $\boldsymbol{T}_{h(i)}^t$ w.r.t. the relative pose increments $\boldsymbol{\xi}$ are, depending on the direction,
\begin{align}
  \frac{\delta \boldsymbol{T}_{h(i)}^t}{\delta \boldsymbol{\xi}(\boldsymbol{T}_{b_{k+1}}^{b_{k}})} &= \Adj_{\boldsymbol{T}_t^{b_k}} \begin{bmatrix}
    \boldsymbol{I}_3 & [\trans(\boldsymbol{T}_{b_{k+1}}^{b_{k}})]_\times \\
    \boldsymbol{0}_{3,3} & \boldsymbol{I}_3
  \end{bmatrix} \\
  \frac{\delta \boldsymbol{T}_{h(i)}^t}{\delta \boldsymbol{\xi}(\boldsymbol{T}_{b_k}^{b_{k+1}})} &= \Adj_{({\boldsymbol{T}_{b_{k-1}}^{t}})^{-1}} \nonumber\\
  &\quad\cdot \begin{bmatrix}
    -\rot(\boldsymbol{T}_{b_k}^{b_{k+1}})^\top & \boldsymbol{0}_{3,3} \\
    \boldsymbol{0}_{3,3} & -\rot(\boldsymbol{T}_{b_k}^{b_{k+1}})^\top
  \end{bmatrix}.
\end{align}
These Jacobians give us the Fisher information of a distribution of the entire state $p(\hat{\boldsymbol{x}} | \boldsymbol{z}_0 ... \boldsymbol{z}_k)$.
In a first step we marginalize out all landmark variables using the Schur complement \cite{sparse-bundle-adjustment}.
This leaves us with a distribution of the relative poses $p(\boldsymbol{T}_0^1, \boldsymbol{T}_1^2, ..., \boldsymbol{T}_{n-1}^n | \boldsymbol{z}_0 ... \boldsymbol{z}_k)$.
After transforming and adding the marginalization prior, we marginalize out all rotation variables, again, using the Schur complement.
This gives us a distribution with only the translation variables $\boldsymbol{t}$ left: $p(\trans(\boldsymbol{T}_0^1), \trans(\boldsymbol{T}_1^2), ..., \trans(\boldsymbol{T}_{n-1}^n) | \boldsymbol{z}_0 ... \boldsymbol{z}_k)$. 
We call the resulting matrix $\boldsymbol{H}_{\boldsymbol{t}}$ the relative translation information matrix. 
Let $\boldsymbol{H}_{\boldsymbol{t}} = \boldsymbol{Q} \diag(\boldsymbol{\lambda}) \boldsymbol{Q}^{-1}$ be the eigenvalue decomposition of $\boldsymbol{H}_{\boldsymbol{t}}$. 
Inspired by Principal Component Analysis, we propose to normalize the inverse square root of the smallest eigenvalue by the average relative translation of the current optimization window: 
\begin{equation}
  \Lambda_{\mathrm{scale drift}} = \left( \sqrt{\min(\boldsymbol{\lambda})} \  \frac{1}{n} \sum_{k=0}^{n-1} | \trans( \boldsymbol{T}_k^{k+1} ) | \right)^{-1}. \label{eq:scale-uncertainty}
\end{equation}
This gives us a metric for estimating the expected relative scale drift, which can be used in decision making. 
A path planner, for example, could decide to fly lower, return to the home point or actively search for an object of known size.

\begin{table*}[!tb]
  \caption{RMS RPE of the longest run on the MADMAX dataset~\cite{madmax}. Calculated with $\Delta = 4\,\textrm{s}$. Below each entry, the maximum relative sequence length ("tracking percentage", see Fig.~\ref{fig:madmax-percent-tracked}.) is given in parentheses. For runs with tracking percentage $>50\%$, the best RMS RPE result is highlighted.}
  \label{tab:madmax-rpe}
  \centering
  \resizebox{\textwidth}{!}{%
  \begin{tabular}{| c | c c c c c c c c c c c c c c c c c c |}
      \hline
      & A-0 & A-1 & A-2 & A-3 & A-4 & A-5 & A-6 & B-0 & B-1 & B-2 & B-3 & B-4 & B-5 & B-6 & B-7 & C-0 & C-1 & C-2 \\
      \hline
      DSO \cite{DSO} & 0.461 & 0.446 & - & 1.484 & 0.353 & 0.570 & 1.063 & 0.287 & 0.473 & 0.545 & 0.307 & 0.690 & 0.412 & 0.807 & 0.642 & 0.260 & 0.359 & 0.423 \\
      & ({\scriptsize{95\%}}) & ({\scriptsize{7\%}})  & ({\scriptsize{0\%}})  & ({\scriptsize{23\%}})  & ({\scriptsize{0\%}}) & ({\scriptsize{16\%}}) & ({\scriptsize{8\%}}) & ({\scriptsize{21\%}}) & ({\scriptsize{97\%}}) & ({\scriptsize{84\%}}) & ({\scriptsize{97\%}}) & ({\scriptsize{20\%}}) & ({\scriptsize{42\%}}) & ({\scriptsize{99\%}}) & ({\scriptsize{99\%}}) & ({\scriptsize{71\%}}) & ({\scriptsize{47\%}}) & ({\scriptsize{27\%}}) \\
      \hline
      ORB-SLAM3 \cite{ORB-SLAM3} & 0.306 & \textbf{0.668} & 0.273 & 0.980 & 1.368 & 1.190 & 1.290 & 0.447 & 0.252 & 0.200 & \textbf{0.222} & \textbf{0.444} & 0.798 & 0.788 & 0.661 & 0.341 & 1.450 & \textbf{0.551} \\
      & ({\scriptsize{92\%}}) & ({\scriptsize{63\%}}) & ({\scriptsize{90\%}}) & ({\scriptsize{55\%}}) & ({\scriptsize{28\%}}) & ({\scriptsize{51\%}}) & ({\scriptsize{44\%}}) & ({\scriptsize{43\%}}) & ({\scriptsize{96\%}}) & ({\scriptsize{94\%}}) & ({\scriptsize{93\%}}) & ({\scriptsize{63\%}}) & ({\scriptsize{97\%}}) & ({\scriptsize{92\%}}) & ({\scriptsize{65\%}}) & ({\scriptsize{98\%}}) & ({\scriptsize{97\%}}) & ({\scriptsize{78\%}})\\
      \hline
      Ours & \textbf{0.278} & 0.673 & \textbf{0.272} & \textbf{0.813} & \textbf{0.949} & \textbf{0.831} & \textbf{1.181} & \textbf{0.659} & \textbf{0.234} & \textbf{0.194} & 0.410 & 1.936 & \textbf{0.685} & \textbf{0.635} & \textbf{0.633} & \textbf{0.258} & \textbf{0.218} & 0.611 \\
      & ({\scriptsize{100\%}}) & ({\scriptsize{100\%}}) & ({\scriptsize{99\%}}) & ({\scriptsize{96\%}}) & ({\scriptsize{65\%}}) & ({\scriptsize{100\%}}) & ({\scriptsize{53\%}}) & ({\scriptsize{78\%}}) & ({\scriptsize{100\%}}) & ({\scriptsize{97\%}}) & ({\scriptsize{100\%}}) & ({\scriptsize{100\%}}) & ({\scriptsize{100\%}}) & ({\scriptsize{100\%}}) & ({\scriptsize{100\%}}) & ({\scriptsize{71\%}}) & ({\scriptsize{65\%}}) & ({\scriptsize{74\%}})\\
      \hline\hline
      & D-0 & D-1 & D-2 & D-3 & D-4 & E-0 & E-1 & E-2 & F-0 & F-1 & F-2 & F-3 & F-4 & F-5 & G-0 & G-1 & G-2 & H-0 \\
      \hline
      DSO \cite{DSO} & 0.368 & 0.395 & 0.166 & 0.128 & 0.226 & 0.493 & 0.441 & 0.261 & 0.180 & 0.193 & 0.178 & 0.222 & 0.178 & 0.518 & 0.547 & 0.182 & 0.203 & 0.817 \\
      & ({\scriptsize{0\%}}) & ({\scriptsize{11\%}}) & ({\scriptsize{13\%}}) & ({\scriptsize{3\%}}) & ({\scriptsize{2\%}}) & ({\scriptsize{7\%}}) & ({\scriptsize{33\%}}) & ({\scriptsize{34\%}}) & ({\scriptsize{97\%}}) & ({\scriptsize{95\%}}) & ({\scriptsize{94\%}}) & ({\scriptsize{18\%}}) & ({\scriptsize{21\%}}) & ({\scriptsize{96\%}}) & ({\scriptsize{70\%}}) & ({\scriptsize{9\%}}) & ({\scriptsize{8\%}}) & ({\scriptsize{96\%}}) \\
      \hline
      ORB-SLAM3 \cite{ORB-SLAM3} & 0.133 & 0.140 & 0.140 & \textbf{0.202} & 0.270 & 0.475 & 1.991 & 0.976 & 0.167 & 0.181 & \textbf{0.165} & 0.705 & 1.690 & 0.517 & 0.334 & 0.276 & 1.380 & 0.348 \\
      & ({\scriptsize{90\%}}) & ({\scriptsize{89\%}}) & ({\scriptsize{92\%}}) & ({\scriptsize{71\%}}) & ({\scriptsize{27\%}}) & ({\scriptsize{58\%}}) & ({\scriptsize{50\%}}) & ({\scriptsize{98\%}}) & ({\scriptsize{89\%}}) & ({\scriptsize{96\%}}) & ({\scriptsize{91\%}}) & ({\scriptsize{91\%}}) & ({\scriptsize{47\%}}) & ({\scriptsize{90\%}}) & ({\scriptsize{49\%}}) & ({\scriptsize{34\%}}) & ({\scriptsize{30\%}}) & ({\scriptsize{56\%}})\\
      \hline
      Ours & \textbf{0.105} & \textbf{0.124} & \textbf{0.131} & 0.216 & \textbf{0.924} & \textbf{0.376} & \textbf{1.783} & \textbf{0.638} & \textbf{0.158} & \textbf{0.180} & 0.173 & \textbf{0.702} & \textbf{1.661} & \textbf{0.494} & \textbf{0.535} & \textbf{0.413} & 0.313 & \textbf{0.291} \\
      & ({\scriptsize{97\%}}) & ({\scriptsize{98\%}}) & ({\scriptsize{99\%}}) & ({\scriptsize{99\%}}) & ({\scriptsize{96\%}}) & ({\scriptsize{93\%}}) & ({\scriptsize{90\%}}) & ({\scriptsize{74\%}}) & ({\scriptsize{100\%}}) & ({\scriptsize{100\%}}) & ({\scriptsize{95\%}}) & ({\scriptsize{96\%}}) & ({\scriptsize{98\%}}) & ({\scriptsize{100\%}}) & ({\scriptsize{100\%}}) & ({\scriptsize{100\%}}) & ({\scriptsize{44\%}}) & ({\scriptsize{100\%}})\\
      \hline
  \end{tabular}}
\end{table*}

To assess the validity of Eq. (\ref{eq:scale-uncertainty}) perfect ground truth is needed.
Thus, we use a photo-realistic simulation (Fig. \ref{fig:simulation-environment}).
In this setup the virtual MAV flies up- and downwards in a spiral with constant speed.
Fig.~\ref{fig:mono-neg-entropy} shows the MAV height, the resulting scale drift as well as $\Lambda_{\mathrm{scale drift}}$. 
To show the effects of accumulated scale drift we include the Relative Pose Error (RPE)~\cite{TUM-RGBD}.
As can be seen, $\Lambda_{\mathrm{scale drift}}$ is a good indicator for the actual scale drift.

%% file: chapters/evaluation.tex

\section{COMPARISON WITH STATE OF THE ART}\label{sec:benchmarks}

We evaluate our method on the MADMAX dataset~\cite{madmax} as it is recorded in an environment similar to the Martian surface. 
Due to its repetitive textures, unstructured terrain and frame drops it is a challenging benchmark dataset relevant to planetary robotic missions.
In Fig.~\ref{fig:madmax-percent-tracked}, we show the percentage of the overall sequence that was tracked without failing (for comparison with results of VINS-Mono and ORB-SLAM2 using both visual and inertial sensor data see~\cite{madmax}).
Note that ORB-SLAM3 and our method can restart after failure and therefore return the longest continuously tracked section whereas DSO does not have such a feature. 
While ORB-SLAM3 fails at random points in time, our method only looses track when frame drops and fast motion coincide. 
In Table~\ref{tab:madmax-rpe}, the Root Mean Squared Relative Pose Error (RMS RPE) is shown, indicating that our method is also more precise.
Given the ground truth pose $\boldsymbol{Q}_t \in SE(3)$ and the estimated pose $\boldsymbol{T}_t \in SE(3)$ at time $t$ we define the RMS RPE for estimated trajectories with unknown scale analogously to~\cite{TUM-RGBD} as 
\begin{align}
    \delta \boldsymbol{Q}_t &= \boldsymbol{Q}_{t-\Delta}^{-1} \boldsymbol{Q}_{t} \qquad
    \delta \boldsymbol{T}_t = \boldsymbol{T}_{t-\Delta}^{-1} \boldsymbol{T}_{t} \\
    s_t &= \frac{| \trans (\delta \boldsymbol{Q}_t) |}{| \trans (\delta \boldsymbol{T}_t) |} \\
    \mathrm{RMS\ RPE} &= \sqrt{ \frac{1}{|\mathcal{T}|} \sum_{t \in \mathcal{T}} | s_t \trans(\delta \boldsymbol{T}_t) - \trans(\delta \boldsymbol{Q}_t) |^2 }.\label{eq:RMS-RPE}
\end{align}
Here, $\mathcal{T}$ is the set of timestamps for which a pose is estimated without those from the first $\Delta$ seconds. 
To get the poses $\boldsymbol{Q}_t$, $\boldsymbol{Q}_{t-\Delta}$ and $\boldsymbol{T}_{t-\Delta}$ we use linear interpolation on the trajectories. 
By calculating the scale factor $s_t$ we eliminate the scale and only measure the error of the estimated relative direction. 
Since the ground truth has a temporal resolution of~$1\,\mathrm{s}$, we computed the values shown in Table~\ref{tab:madmax-rpe} with $\Delta = 4\,\textrm{s}$. 
Not visible in the RMS RPE is that ORB-SLAM3, by performing global bundle adjustment, shows less scale drift.
Of course, our approach could also benefit from global optimization of all keyframes.

%% file: chapters/conclusions.tex

\section{CONCLUSIONS}

We presented a monocular odometry that is especially robust in environments relevant for planetary exploration.
We achieved this by using KLT tracking with features detected at different image pyramid levels and by refining the keyframe selection strategy of~\cite{arbitrary-camera-slam}. 
We detailed our map initialization strategy since for monocular setups, special care has to be taken at the start of a new trajectory.
By allowing landmarks with inverse distance $d=0$, our system can handle rotation-only motion sequences. 
Compared to other methods our system shows significantly less tracking failures and high local accuracy on the MADMAX dataset~\cite{madmax}. 
In Eq.~(\ref{eq:scale-uncertainty}), we proposed a quantity that allows for the detection of scale drift which is crucial to overall system robustness. 
It is straight forward to apply a similar criterion to detect when a stereo setup degrades to a monocular one. 
This happens during flights at high altitude, as the stereo baseline cannot constrain the scene scale well anymore. 
We will further explore the practical application of our findings during the ARCHES mission~\cite{arches}.